\documentclass[letterpaper, 10 pt, conference]{ieeeconf}  

\IEEEoverridecommandlockouts                              

\usepackage{xcolor}         
\usepackage{makecell}
\usepackage{amsmath}
\usepackage{graphicx}
\usepackage{multirow}
\usepackage{booktabs}
\usepackage{colortbl}
\usepackage{url}
\usepackage{graphics} 
\usepackage{epsfig} 
\usepackage{mathptmx} 
\usepackage{times} 
\usepackage{amsmath} 
\usepackage{amssymb}  

\usepackage{censor}

\title{StereoSplat+: Feed-Forward Stereo Gaussian Splatting with Diffusion-Assisted Progressive Inference}

\author{Zihua Liu$^{1}$ and Masatoshi Okutomi.$^{1}$
\thanks{$^{1}$ Department of systems and Control Engineering, Institute of Science Tokyo, Japan.
{\tt\small \{zliu,mxo\}@ok.sc.e.titech.ac.jp}}
}

\begin{document}

\maketitle
\thispagestyle{empty}
\pagestyle{empty}


\begin{abstract}
\label{sec:abstract}
Recent advances in 3D Gaussian Splatting (3DGS) have enabled high-quality, render-ready scene representations for novel-view synthesis. However, most existing 3DGS pipelines rely on multi-view observations (or non-causal access to future frames) to achieve sufficient coverage, which is often unavailable in on-device robotics and AR settings where sensing is restricted to a single stereo rig. Recovering a high-quality 3DGS scene from one stereo observation, therefore, remains challenging due to occlusions, limited field of view, and missing geometry.
We present StereoSplat+, a diffusion-enhanced feed-forward framework that enables causal reconstruction from a single stereo pair. Our method builds on two key components. First, we propose \textit{StereoSplat}, an input-invariant feed-forward 3D Gaussian estimator that takes a variable number of posed stereo pairs as input and predicts high-quality 3D Gaussians. StereoSplat fuses complementary geometry cues via a cost-volume branch and a triplane-based 3D volume branch, and leverages continuous pose encoding to generalize across view counts and camera configurations. Second, since multiple posed stereo pairs are typically unavailable at inference time, we introduce a diffusion-enhanced one-shot progressive inference scheme called \textit{StereoSplat+}: starting from one stereo pair, we render novel stereo views from the predicted 3DGS, refine them with a one-step diffusion enhancer, and feed them back as additional inputs to update the 3DGS. Experiments on the KITTI-360 dataset show that StereoSplat+ improves novel-view rendering quality and geometry accuracy, especially in occluded regions and under strong view extrapolation, outperforming recent feed-forward 3DGS baselines.

\end{abstract}
\section{Introduction} 
\label{sec:introduction}

Feed-forward 3D Gaussian Splatting (3DGS)~\cite{AnySplat,OmniScene,DepthSplat,pixelsplat,MVSPlaT} enables real-time and generalizable scene reconstruction by directly predicting a set of view-renderable Gaussians from multi-view inputs. However, in many practical robotics and on-device AR settings, the input is often restricted to a single stereo pair. With such limited coverage and field of view, distant surfaces and regions occluded in the stereo observations are only weakly constrained, causing existing feed-forward 3DGS pipelines to struggle with reliable geometry coverage and photorealistic novel-view synthesis. Incorporating more views can mitigate these issues, but it increases latency and typically introduces non-causal requirements (e.g., access to future frames or accurate multi-view poses), which are incompatible with real-time deployment.

\begin{figure}
\centering
\includegraphics[width=1.0\linewidth]{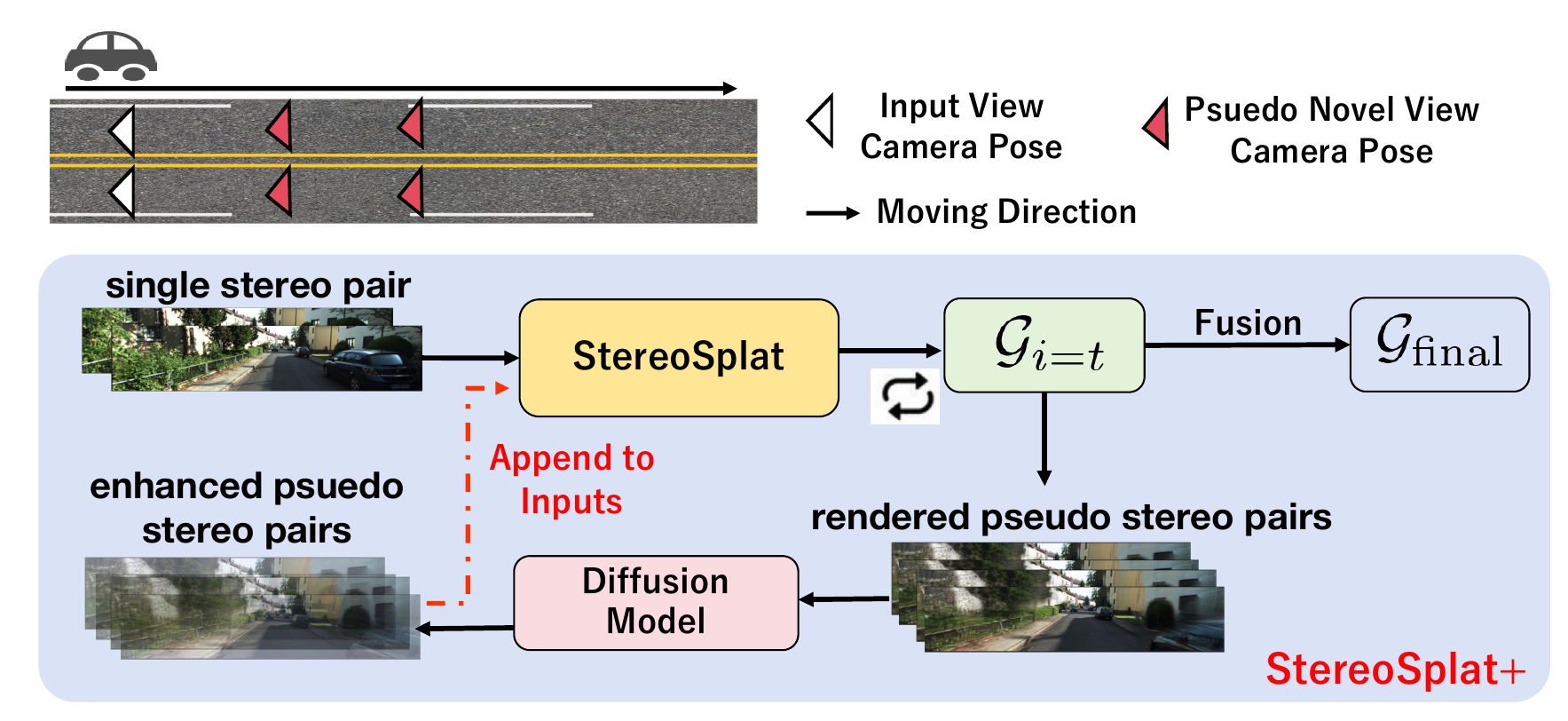}
\caption{The pipeline of our proposed \textit{StereoSplat+}. StereoSplat+ performs progressive inference: starting from one stereo pair, it estimates an initial 3DGS, renders novel stereo views, refines them with a one-step diffusion prior, and re-injects the enhanced views as pseudo inputs to update the 3D Gaussians.}
\label{fig:pipeline_overall}
\end{figure}
To tackle this issue, we introduce \textit{StereoSplat+}, an input-invariant, feed-forward 3DGS framework for a single stereo pair input with \emph{diffusion-enhanced one-shot progressive inference}. As shown in Figure~\ref{fig:pipeline_overall}, compared with conventional feed-forward 3DGS estimator,our key idea is to couple a view-count-agnostic 3DGS predictor with a single-step diffusion enhancer for progressive inference:
Starting from one stereo pair, we predict a provisional 3D Gaussian set, render novel stereo views, enhance the rendered images with a one-step diffusion model, and re-inject the enhanced views as \emph{pseudo views} to provide additional geometric cues for new Gaussian estimation. The final result is the confidence-based fusion of the single-stereo Gaussians and pseudo-multi-stereo Gaussians.
Importantly, we perform \emph{exactly one} render--enhance--reinject round, which densifies multi-view evidence without breaking the feed-forward, real-time nature of the pipeline.

At the core of \textit{StereoSplat+} is a dual-branch architecture that combines a classical \textit{cost-volume} branch with a lightweight \textit{3D volume} branch. The cost-volume branch preserves accurate stereo cues, while the 3D branch aggregates geometric evidence to better handle occlusions and long-baseline reasoning. Both branches are jointly optimized to predict depth and 3D Gaussian parameters, eliminating the need for an external depth estimator. To support variable input stereo pairs, we employ continuous sinusoidal pose encodings together with stochastic view sampling, which randomly varies both the number and locations of input views during training. To further improve robustness, sampled GT stereo pairs are randomly replaced by rendered pseudo stereo pairs with probability $p$, encouraging the model to learn from both clean and degraded observations. During inference, \textit{StereoSplat} first predicts an initial 3DGS and renders novel stereo pairs as pseudo-observations. Following~\cite{difix3d+}, these pseudo views are enhanced by a one-step diffusion model before being re-injected into the network, producing a refined 3DGS.  

We evaluate \textit{StereoSplat+} on KITTI-360~\cite{kitti360} for both rendered RGB and depth. Starting from a single stereo pair, our method outperforms prior feed-forward methods, and diffusion-enhanced progressive inference further improves photometric fidelity and geometric in occluded and out-of-frame regions—while preserving the efficiency of a feed-forward pipeline.

\noindent Our contributions are summarized as follows:

\begin{itemize}
    \item We propose \textit{StereoSplat+}, a progressive, input-invariant feed-forward 3D Gaussian Splatting framework tailored to stereo inputs.
    \item We design a dual-branch predictor that fuses cost-volume cues with a lightweight 3D volume branch to jointly estimate depth and Gaussian parameters.
    \item We introduce a one-step diffusion enhancer that improves rendered pseudo novel views and re-injects them as additional inputs for one-shot progressive refinement.
    \item We develop a robust training strategy that combines continuous pose encodings with stochastic view subsampling, making the model invariant to both the number of input stereo pairs and their camera placements.
    \item Experiments on KITTI-360~\cite{kitti360} show consistent gains over feed-forward 3DGS baselines on novel-view and depth metrics, with clear qualitative improvements in weakly constrained regions.
\end{itemize}

\section{Related Work}
\label{sec:related_works}

\subsection{Feed-Forward 3D Gaussain Splatting}
\label{sec:related_works:feedforward3dgs}
Feed-forward 3D Gaussian Splatting predicts a scene’s 3D Gaussians from a small set of posed images in a single forward pass~\cite{Kerbl2023GaussianSplatting}, avoiding slow per-scene optimization. Recent methods illustrate this trend: PixelSplat~\cite{pixelsplat} uses a multi-view epipolar Transformer to recover metric scale and infer Gaussian parameters from dense probability maps, while MVSplat~\cite{MVSPlaT} builds a stereo-style cost volume, which is often used in stereo matching methods~\cite{GOAT,NiNet}, to jointly estimate depth and Gaussian attributes. DepthSplat~\cite{DepthSplat} further strengthens this pipeline by tightly coupling depth prediction with 3DGS estimation, allowing the two tasks to benefit each other. Object-centric variants such as LatentSplat~\cite{latentsplat} combine variational Gaussians with a lightweight generative decoder for novel-view synthesis. However, estimating 3DGS from a single stereo pair remains difficult: occlusions, low-texture regions, and repetitive patterns often lead to splat drift, floaters, and over-smoothed geometry.

\subsection{Diffusion-Incorporated 3D Gaussian Splatting}
\label{sec:diff_3dgs}
Diffusion-incorporated 3DGS falls into three concise categories: (1) Text/Image-to-3D pipelines~\cite{dreamgaussian,dreamergaussianpro} that use 2D diffusion priors to directly synthesize render-ready Gaussian scenes. (2) Diffusion-guided repair/inpainting that refines or completes sparse-view 3DGS by enhancing rendered views and feeding them back as constraints~\cite{gsfix3d,difix3d+,ri3d}. (3) Diffusion-assisted supervision~\cite{cascade_zero123,viewcrafter}, where diffusion produces view-consistent auxiliary signals to regularize geometry when multi-view evidence is weak. Together, these works show diffusion can create, stabilize, and supervise Gaussian reconstructions. Our method is aligned with this direction, but remains feed-forward and lightweight: we use diffusion only to strengthen under-constrained regions, without heavy per-scene optimization.
\section{Method}
\label{sec:proposed_method}

In this section, we present the proposed \textit{StereoSplat+} pipeline. We first outline the overall architecture of StereoSplat+ in Subsection~\ref{sec:proposed_method:overall_architecture}. Next, Subsection~\ref{sec:proposed_method:input_invariant_stereosplat} details the input-invariant StereoSplat, including the network architecture and training strategy. Finally, Subsection~\ref{sec:proposed_method:progressive_3dgs} describes our progressive, input-invariant feed-forward 3DGS pipeline, StereoSplat+ with a diffusion model.


\begin{figure*}[!th]
\centering
\includegraphics[width=1.0\linewidth]{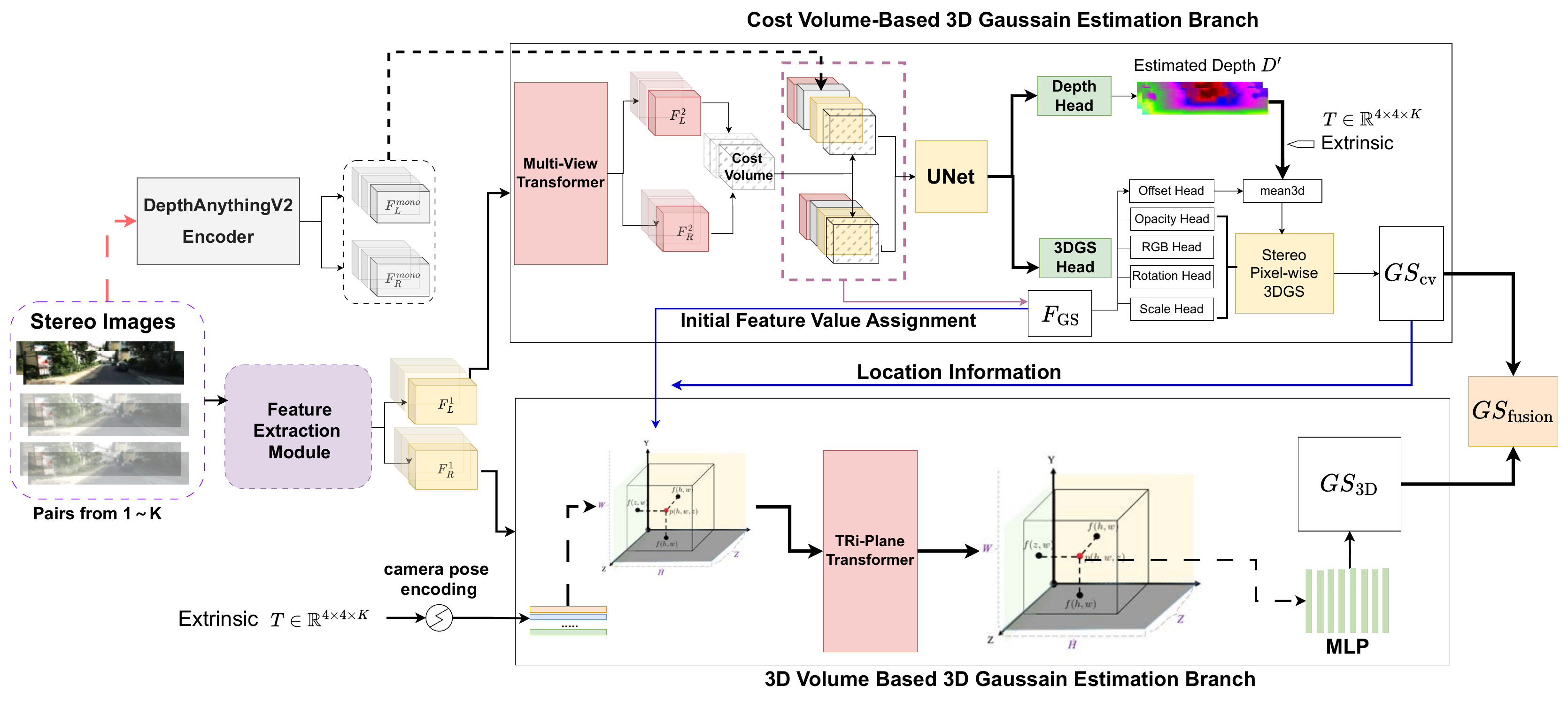}
\caption{Overall architecture of \textit{StereoSplat}. Given a variable number of stereo pairs (1 to $K$), we extract image and geometry features using a pre-trained Depth-Anything V2 backbone, and predict 3D Gaussians via two complementary heads: a cost-volume branch and a triplane Transformer branch that reasons in 3D space. Following OmniScene~\cite{OmniScene}, cross-branch interaction is enabled by feature-value assignment guided by Gaussian locations estimated by the cost-volume branch. The fused 3DGS is finally rendered for novel-view synthesis.}

\label{fig:stereosplat}
\end{figure*}

\subsection{The Overall Pipeline}
\label{sec:proposed_method:overall_architecture}
Estimating a reliable 3D Gaussian representation from only \emph{single} stereo pair is challenging: the field of view is limited, many regions are occluded, and evidence outside the cameras’ view is missing. Inspired by diffusion-assisted 3DGS reconstruction (e.g.,~\cite{difix3d+}), we use diffusion \emph{not} to directly generate novel views, but as a fast one-step prior that refines rendered pseudo views with more view-consistent cues. This compensates for the limited stereo input and strengthens geometric evidence during inference. As shown in Fig.~\ref{fig:pipeline_overall}, unlike prior methods that directly consume multi-view posed images and estimate 3DGS in a single pass, our proposed \textit{StereoSplat+} pipeline starts from a \emph{single} rectified stereo pair. We first run \textit{StereoSplat}—our input-invariant 3DGS estimator—to obtain an initial Gaussian set $\mathcal{G}^{(0)}$ (at time $t$). We then render novel stereo views from $\mathcal{G}^{(0)}$ as pseudo inputs and enhance them with a \emph{one-step diffusion enhancer} to improve structure and suppress artifacts. Finally, we append the enhanced pseudo views to the original inputs and re-run \textit{StereoSplat} to predict a refined Gaussian set $\mathcal{G}^{(1)}$ with richer geometric evidence.

To realize this pipeline, we hinge on \textbf{two} main components:
(1) An input-invariant feed-forward stereo 3DGS estimator \textit{StereoSplat}: it accepts a variable number of views and remains robust to their relative camera pose, while still producing high-fidelity renders from minimal input.  To further improve robustness under progressive inference, we additionally employ pseudo-view mixed training by randomly replacing sampled GT stereo pairs with rendered pseudo-stereo pairs during training. We will introduce our proposed input-invariant feed-forward 3DGS network architecture and training scheme in Subsection~\ref{sec:proposed_method:input_invariant_stereosplat}. 
(2) A one-step diffusion enhancer that reliably improves the low-quality renders produced under sparse views, supplying sharper, more view-consistent images that can be appended as additional inputs. More details can be seen at Subsection~\ref{sec:proposed_method:progressive_3dgs}. After integrating the input-invariant 3DGS estimator with the diffusion enhancer, our pipeline can emulate \emph{pseudo multi-view} inputs by one-shot progressive inference, even when only a single stereo pair is available.

\subsection{Input-Invariant StereoSplat}
\label{sec:proposed_method:input_invariant_stereosplat}
In this section, we introduce the network architecture of our input-invariant, feed-forward 3DGS estimator: \textit{StereoSplat}. As shown in Figure.~\ref{fig:stereosplat}, we begin with 1 to $K$ stereo image pairs $I = \{I^{l}_{i},I^{r}_{i}\}^{k}_{i=1}$ and their corresponding camera pose $T=\{T^{l}_{i},T^{r}_{i}\}^{k}_{i=1}$ and camera intrinsics $K=\{K^{l}_{i},K^{r}_{i}\}^{k}_{i=1}$ as input, we first use a shared CNN-based feature extraction backbone to extract image features together with geometry feature with a pre-trained Depth-Anything v2 model~\cite{DepthAnythingV2}, the geometry feature and the image feature are further used to estimate the 3D Gaussain attributes. Then we adopt a two-branch design: a cost-volume branch that preserves metric stereo cues and a triplane transformer branch that aggregates 3D evidence in Euclidean space. The two branches are coupled via feature-value assignment guided by the Gaussian locations estimated by the cost-volume branch, yielding a fused 3D Gaussian set for novel-view synthesis. Our design is inspired by DepthSplat~\cite{DepthSplat} and OmniScene~\cite{OmniScene}. DepthSplat constructs a cost volume from stereo images and regresses 3DGS attributes from matching signals; however, lacking explicit 3D aggregation, its improvements in unmatched regions are limited. OmniScene introduces a pixel/volume fusion mechanism for better 3DGS estimation, but the pixel branch assumes precomputed depth (e.g., Metric3D-v2~\cite{Metric3D}), placing strong robustness demands on metric monocular depth estimation, which is usually less reliable in practice. In contrast, StereoSplat performs joint estimation of depth and 3D Gaussians: we retain the stereo-friendly cost-volume structure of DepthSplat while introducing a triplane transformer to aggregate in 3D space, which better suits our sparse-stereo input setting.

\subsubsection{Cost Volume-Based 3D Gaussian Estimation Branch.}

\textbf{Multi-View Cost Volume Construction.}
Similar to DepthSplat~\cite{DepthSplat}, we construct a stereo cost volume by combining geometry features
$F_L^{\text{mono}}, F_R^{\text{mono}}$ from a pre-trained Depth-Anything V2 backbone and aggregated image features
$F_L^{\text{i'}}, F_R^{\text{i'}}$.
The aggregated features are produced by a multi-view Transformer built with Swin Transformer blocks~\cite{SwinTransformer},
using both self/cross-attention across the left and right views.
We concatenate them to form per-view features:
\begin{equation}
\mathbf{F}_v(\mathbf{x})
=
\Big[
F_v^{\text{mono}}(\mathbf{x}) \,\Vert\,
F_v^{\text{i'}}(\mathbf{x})
\Big],
\quad v\in\{L,R\}.
\end{equation}
To further support a variable number of stereo pairs and make the pipeline input-invariant, we build a cost volume per reference view.
Assume $K$ stereo pairs (in total $2K$ views). Let $I_i$ denote the $i$-th input image with camera pose $T_i \in SE(3)$.
We define a pose-based distance
\begin{equation}
d(i,j)=\delta(T_i, T_j), \qquad j\in\{1,\dots,2K\}\setminus\{i\},
\label{eq:dist_ours}
\end{equation}
where $\delta(\cdot,\cdot)$ measures the distance between the translation components of two camera poses.
For each reference view $i$, we select its $N$ nearest views:
\begin{equation}
M_i=\operatorname*{arg\,TopN}_{\,j\neq i}\big(-d(i,j)\big),
\qquad N<2K-1,
\label{eq:matchset_ours}
\end{equation}
And then we build a pairwise correlation cost volume with each selected view:
\begin{equation}
\mathbf{C}_{i\leftrightarrow j}=\Psi\!\big(I_i,I_j;\,T_i,T_j\big),
\qquad j\in M_i,
\label{eq:pair_cost_ours}
\end{equation}
and average them to obtain an input-invariant cost volume:
\begin{equation}
\mathbf{C}_i=\frac{1}{|M_i|}\sum_{j\in M_i}\mathbf{C}_{i\leftrightarrow j}.
\label{eq:avg_cost_ours}
\end{equation}
Here, $\Psi(\cdot)$ denotes the warping-based feature correlation (cost-volume) builder.
In our experiments, we use 192 disparity candidates. For the depth estimation, We attach a lightweight 2D U-Net depth head to predict a per-view depth map $D'_i$ from the stacked multi-view inputs.

\noindent \textbf{Gaussian Parameter Prediction.}
Following DepthSplat~\cite{DepthSplat}, we predict 3D Gaussian parameters (center $\mu$, opacity $\alpha$, covariance $\Sigma$,
and color $c$) with lightweight heads.
For each pixel $(u,v)$ in view $i$, we back-project the predicted depth to the camera frame and learn a 3D offset
$\Delta \mathbf{p}_i(u,v)$ to compensate for the mismatch between raw points and Gaussian means:
\begin{equation}
\mu_i(u,v)
=
T_i\Big(
D'_i(u,v)\,K_i^{-1}[u, v, 1]^{\top}
+
\Delta \mathbf{p}_i(u,v)
\Big).
\end{equation}
For opacity, we derive a matching confidence by applying a softmax over the cost volume and map it to $\alpha$ using two convolution layers followed by a sigmoid activation. Color is represented by spherical harmonic (SH) coefficients; in practice, we adopt degree-0 SH, resulting in a view-independent RGB color $c$ for each Gaussian. Additionally, a confidence head with a sigmoid activation predicts a confidence attribute for each Gaussian.

\subsubsection{3D Volume-Based 3D Gaussian Estimation Branch.}
The cost-volume branch predicts pixel-aligned Gaussians $G_{\text{cv}}$ from stereo depth cues, but it inherits stereo matching failures in ill-conditioned regions (e.g., occlusions and weak texture), which often degrade reconstruction near disocclusions. To complement it, we introduce a 3D volume branch that aggregates evidence directly in 3D. Inspired by OmniScene~\cite{OmniScene}, we adopt a Triplane Transformer~\cite{triplane_transformer} that factorizes an $H \times W \times Z$ volume into three axis-aligned planes ($HW$, $HZ$, $WZ$), improving robustness when image-space correspondences are unreliable. The triplane is conditioned on multi-view image features $F_i^{l}, F_i^{r}$, cost-volume Gaussian features $F_{\text{GS}}$, and camera-pose encodings that disambiguate feature provenance across views.

\noindent \textbf{Sinusoidal Camera Pose Positional Encoding.}
Different from the Tri-plane transformer proposed in \cite{OmniScene}, to support variable view counts, we replace the fixed-size learnable camera embeddings with a continuous sinusoidal encoding of the camera pose. Given a camera-to-world pose
$T_i=\begin{bmatrix}R_i & t_i\\ 0 & 1\end{bmatrix}\in SE(3)$, we normalize translation and map rotation to $\mathbb{R}^3$ using the $SO(3)$ log map:
\begin{gather}
\phi(x)=\big[\sin(\omega_1x),\cos(\omega_1x),\ldots,\sin(\omega_Mx),\cos(\omega_Mx)\big],\\
\omega_m=2\pi b^{-\frac{m-1}{M-1}},\qquad
x_i=\begin{bmatrix}t_i/s_t\\ \operatorname{Log}(R_i)/s_r\end{bmatrix},\\
\operatorname{PE}(T_i)=
\operatorname{pad}_D\!\left(
\big[
\phi(x_{i,1})\Vert\cdots\Vert\phi(x_{i,C})
\Vert\gamma x_i
\big]
\right).
\end{gather}
Here, $s_t,s_r>0$ scale translation and rotation to $\mathcal{O}(1)$; $\gamma\in\{0,1\}$ optionally appends the raw normalized pose; and $\operatorname{pad}_D(\cdot)$ zero-pads or truncates to the target dimension $D$ (e.g., $D{=}128$). Together with the stochastic view subsampling proposed in Subsection~\ref{method:training:stochastic_stereo_view}, this encoding yields an input-invariant conditioning over both input frame number and relative camera poses.

\noindent \textbf{Triplane Transformer for 3D Feature Aggregation.}
We maintain learnable plane features $\{\mathbf{Q}_{HW},\mathbf{Q}_{HZ},\mathbf{Q}_{WZ}\}$ and update them via alternating cross-image and cross-plane attention, following OmniScene~\cite{OmniScene}. For a query on a plane (take $HW$ as an example), we expand it along $Z$ into sparse pillar points, project them into each input view using $T_i$, and sample view-conditioned 2D features (with $\operatorname{PE}(T_i)$). Cross-image deformable attention (CIDA) aggregates multi-view evidence into the plane query. To propagate context when pillars are occluded or outside the frustum, we apply cross-plane deformable attention (CPDA) by orthogonally projecting pillar points onto the other two planes, enabling lightweight information exchange across planes. One update step is summarized as
\begin{equation}
\begin{aligned}
\underbrace{\mathbf{q}_{HW} \leftarrow \mathrm{DA}\!\Big(\mathbf{q}_{HW},\,\mathrm{Ref}^{2D},\,\{\mathbf{F}_i,\operatorname{PE}(T_i)\}\Big)}_{\text{CIDA}}
\;\Rightarrow\;\\
\underbrace{\mathbf{q}_{HW} \leftarrow \mathrm{DA}\!\Big(\mathbf{q}_{HW},\,\mathrm{Ref}^{3D},\,\{\mathbf{Q}_{HW},\mathbf{Q}_{HZ},\mathbf{Q}_{WZ}\}\Big)}_{\text{CPDA}} .
\end{aligned}
\end{equation}
By performing cross-view and cross-plane attention, the triplane features can effectively model the 3D space and its underlying correlations, providing an explicit representation for volume-wise 3DGS estimation.

\noindent \textbf{Projection-Guided Fusion from the Cost-Volume Branch.}
To explicitly inject metric geometry and improve coverage in occluded/truncated regions, we fuse projected features from the cost-volume branch into the triplane. Concretely, we project Gaussians $G_{\text{cv}}$ and their features $F_{\text{GS}}$ onto each plane using their estimated locations, average-pool features that map to the same plane query, adapt them with a linear layer, and add them to the corresponding plane queries (discarding samples outside the $H \times W \times Z$ volume). The final Gaussian set is formed as $G_{\text{fusion}} = G_{\text{cv}} \cup G_{\text{volume}}.$

\noindent \textbf{Gaussian Parameter Prediction for the 3D Volume Branch.}
For each voxel at $(h,w,z)$ (the \textbf{\textcolor{red}{red}} dot in Figure~\ref{fig:stereosplat}), we bilinearly sample features from the three planes ($HW$, $HZ$, and $WZ$) and average them to obtain an aggregated voxel feature $f_{h,w,z}$. A lightweight predictor then decodes $f_{h,w,z}$ into $V$ voxel-wise Gaussians with parameters $(\alpha,\mu,s,q,c,\mathrm{conf})$, where $\alpha$ denotes opacity, $\mu$ the Gaussian center, $s$ the scale, $q$ the rotation (quaternion), $c$ the color represented by spherical harmonics (SH), and $\mathrm{conf}$ the confidence score.

\subsubsection{Stochastic Stereo View Subsampling for Input-Invariant StereoSplat Training}
\label{method:training:stochastic_stereo_view}
To make \textit{StereoSplat} invariant to both the \emph{number} and \emph{placement} of input stereo views, we adopt stochastic view subsampling within each temporal bin
$B=\{(I_i^L, I_i^R, T_i)\}_{i=1}^{K}$.
At each training iteration, we first sample the number of input stereo pairs
$m \sim \mathcal{U}\{1,\ldots, K\}$,
and then uniformly sample a subset $S \subset B$ with $|S|=m$ as the input.
This introduces randomness in both view count and view positions, encouraging view-agnostic 3D Gaussian estimation. Crucially, our 3D-volume branch aggregates evidence in \emph{scene space} rather than \emph{view space}; thus, stochastic subsampling forces the network to explain consistent geometry under different randomly sampled input sets.
In practice, the predicted Gaussians remain stable under subsampling and typically improve as more stereo pairs are provided.

\subsubsection{Training Loss}
For each supervised view $i$, let $\hat{I}_i^{b}$, $\hat{D}_i^{b}$, and $\hat{C}_i^{b}$ denote the rendered RGB image, depth map, and confidence map from the branch
$b \in \{\text{cv},\text{vol},\text{fuse}\}$, and let $I_i$ be the ground-truth image.
Let $D_i^\dagger$ be the target depth constructed from sparse LiDAR together with pseudo-dense depth predicted by a pre-trained NMRFStereo model~\cite{NMRFStereo}. A validity mask $M_i$ indicates pixels supported by LiDAR. To supervise the confidence, we construct a self-supervised confidence target \(C_i^\dagger=\exp(-\beta E_i)\) to supervise the confidence attribute of each 3D Gaussian, where $E_i$ denotes the detached photometric reconstruction error. The confidence branch is supervised by a standard regression loss $L_{\text{conf}}^{b}$. And the other losses are described as follows:
\begin{align}
L_{\text{rec}}^{b}
&= \frac{\lambda_{1}}{N}\sum_i \big\|\hat{I}_i^{b}-I_i\big\|_2^2
 \;+\; \frac{\lambda_{2}}{N}\sum_i
\mathrm{LPIPS}\!\left(\hat{I}_i^{b},I_i\right),
\\[2pt]
L_{\text{D-est}}
&= \frac{1}{N}\sum_i
\big\| D'_i-D_i^\dagger \big\|_{1,M_i},
\\[2pt]
L_{\text{D-rend}}^{b}
&= \frac{1}{N}\sum_i
\big\| \hat{D}_i^{b}-D_i^\dagger \big\|_{1,M_i}.
\end{align}

Here, $N$ is the number of supervised views in the current batch, and
$\|X\|_{1,M}=\frac{\sum M\cdot|X|}{\sum M}$ denotes the masked $\ell_1$ norm.

\noindent
The total objective is
\begin{align}
L
=\sum_{}
\Big(
\lambda_{\text{1}}L_{\text{rec}}
+\lambda_{\text{2}}L_{\text{D-rend}}
+\lambda_{\text{3}}L_{\text{conf}}
\Big)
+\lambda_{\text{4}}L_{\text{D-est}}.
\label{chapter6:eq:loss_function}
\end{align}

\subsection{Progressive Input-Invariant Feed-Forward 3DGS with Diffusion Models}
\label{sec:proposed_method:progressive_3dgs}
As shown in Figure~\ref{fig:pipeline_overall}, we propose \textit{StereoSplat+}, a progressive inference framework that alleviates the limited scene coverage of a single stereo pair. Starting from the input stereo pair, we estimate an initial 3D Gaussian representation $G_{\textbf{base}}$ together with a confidence attribute $C$, which enables confidence-aware rendering. The rendered novel stereo views $I_{\textbf{base}}$ are subsequently refined using a one-step diffusion enhancer~\cite{SD_Turbo} and fed back as pseudo multi-view inputs to re-estimate the 3D Gaussian representation $G_{\textbf{plus}}$ and its confidence maps. The final predictions are obtained through confidence-guided fusion, which adaptively combines the renderings from the two stages according to their estimated confidence. During training, the pseudo views may exhibit appearance and geometric discrepancies with respect to the ground-truth views. To improve robustness against pseudo-view inputs, we randomly replace sampled ground-truth stereo pairs with pseudo-stereo pairs rendered by the current-stage \textit{StereoSplat} model during training. This reduces the training--inference distribution gap and improves the stability of progressive inference.

\subsection{Training a One-Step Diffusion Enhancer for Image Refinement.} To further improve novel stereo rendering quality, we fine-tune the revised SD-Turbo–based one-step enhancer from DIFIX3D+~\cite{difix3d+} using training pairs generated by our input-invariant StereoSplat. For each bin (Subsection~\ref{sec:exp:implementation_details}), we select the first left image as the reference view, randomly sample the number of input stereo pairs $K$, estimate 3DGS with the pre-trained model, and render novel views. The rendered images and their corresponding ground-truth views form render $\rightarrow$ GT training pairs for fine-tuning. This enables the enhancer to robustly refine novel views across varying numbers of input stereo pairs $K$.
\section{Experiments}
\label{sec:experiments}

\begin{figure}[!t]
\centering
\includegraphics[width=1.0\linewidth]{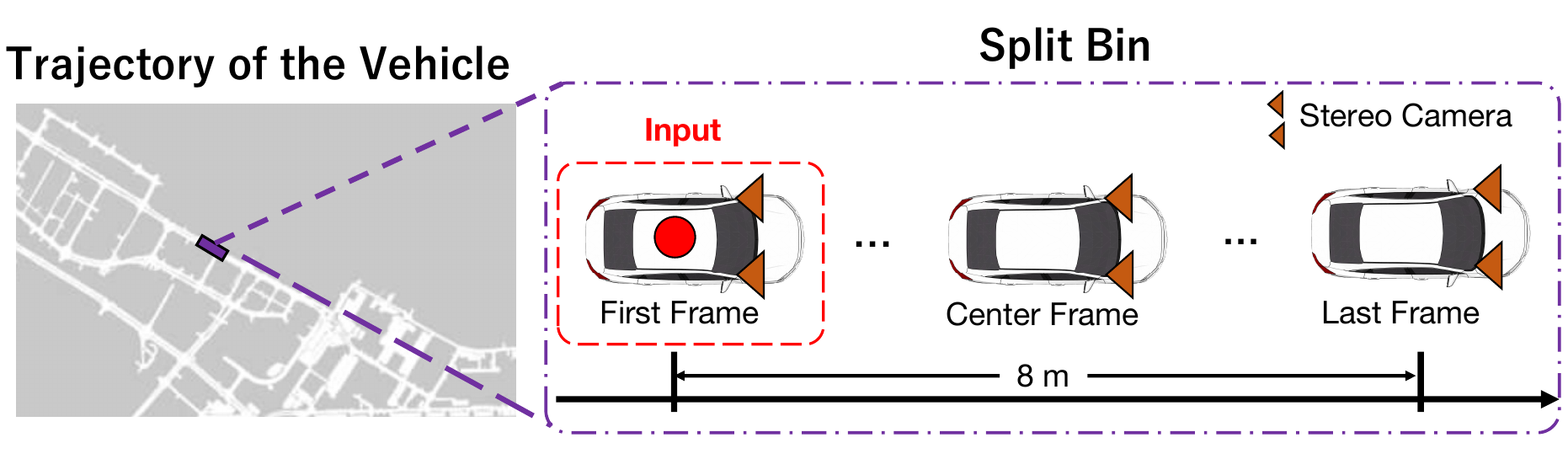}
\caption{Bin partitioning on KITTI-360~\cite{kitti360} dataset. Each sequence is split into bins along the trajectory. In each bin, the first-frame stereo pair is taken as the reference and its LiDAR frame defines the world origin; all target camera poses are expressed relative to this origin. The first stereo pair is used as input, while the remaining frames in the bin provide target novel views for supervision and evaluation.}
\label{fig:dataset_configuration}
\end{figure}

\begin{table*}[!t]
\centering
\setlength{\tabcolsep}{1.8 mm}
\renewcommand{\arraystretch}{1.3} 

\caption{\textbf{Ablation on KITTI-360.} In-bin PSNR/SSIM.
$\text{Branch}_{\text{CV}}$: cost-volume only; $\text{Branch}_{\text{3D}}$: triplane/3D-only.
\textit{One-shot progressive}: predict a provisional 3DGS, render novel views once, optionally enhance with diffusion, then re-predict the final 3DGS from the augmented views.
All ablations use a \textbf{single} real input stereo pair.
Best are reported in \textcolor{red}{\textbf{bold}}.}

\begin{tabular}{ccccccccc}
\hline
\multicolumn{1}{c|}{\multirow{2}{*}{\textbf{Method}}} &
\multicolumn{1}{c|}{\multirow{2}{*}{\textbf{\begin{tabular}[c]{@{}c@{}}Cost Volume\\ Branch\end{tabular}}}} &
\multicolumn{1}{c|}{\multirow{2}{*}{\textbf{\begin{tabular}[c]{@{}c@{}}3D Volume\\ Branch\end{tabular}}}} &
\multicolumn{1}{c|}{\multirow{2}{*}{\textbf{\begin{tabular}[c]{@{}c@{}}Stochastic View\\ Subsampling\end{tabular}}}} &
\multicolumn{1}{c|}{\multirow{2}{*}{\textbf{\begin{tabular}[c]{@{}c@{}}Use\\ Diffusion\end{tabular}}}} &
\multicolumn{1}{c|}{\textbf{\begin{tabular}[c]{@{}c@{}}First\\ Frame\end{tabular}}} &
\multicolumn{1}{c|}{\textbf{\begin{tabular}[c]{@{}c@{}}Center\\ Frame\end{tabular}}} &
\multicolumn{1}{c|}{\textbf{\begin{tabular}[c]{@{}c@{}}Last\\ Frame\end{tabular}}} &
\textbf{\begin{tabular}[c]{@{}c@{}}Bin\\ Average\end{tabular}} \\ \cline{6-9}
\multicolumn{1}{c|}{} &
\multicolumn{1}{c|}{} &
\multicolumn{1}{c|}{} &
\multicolumn{1}{c|}{} &
\multicolumn{1}{c|}{} &
\multicolumn{1}{c|}{\textbf{PSNR/SSIM}} &
\multicolumn{1}{c|}{\textbf{PSNR/SSIM}} &
\multicolumn{1}{c|}{\textbf{PSNR/SSIM}} &
\textbf{PSNR/SSIM} \\ \hline

\multicolumn{1}{c|}{$\text{Branch}_{\text{CV}}$} &
\multicolumn{1}{c|}{\checkmark} &
\multicolumn{1}{c|}{} &
\multicolumn{1}{c|}{} &
\multicolumn{1}{c|}{} &
\multicolumn{1}{c|}{25.20/0.89} &
\multicolumn{1}{c|}{19.27/0.66} &
\multicolumn{1}{c|}{17.21/0.56} &
20.12/0.69 \\ \hline

\multicolumn{1}{c|}{$\text{Branch}_{\text{3D}}$} &
\multicolumn{1}{c|}{} &
\multicolumn{1}{c|}{\checkmark} &
\multicolumn{1}{c|}{} &
\multicolumn{1}{c|}{} &
\multicolumn{1}{c|}{19.83/0.59} &
\multicolumn{1}{c|}{18.31/0.53} &
\multicolumn{1}{c|}{17.15/0.49} &
18.43/0.54 \\ \hline

\multicolumn{1}{c|}{\begin{tabular}[c]{@{}c@{}}$\text{Branch}_{\text{CV}}$ +\\ $\text{Branch}_{\text{3D}}$\end{tabular}} &
\multicolumn{1}{c|}{\checkmark} &
\multicolumn{1}{c|}{\checkmark} &
\multicolumn{1}{c|}{} &
\multicolumn{1}{c|}{} &
\multicolumn{1}{c|}{24.24/0.87} &
\multicolumn{1}{c|}{20.03/0.67} &
\multicolumn{1}{c|}{18.32/0.57} &
20.53/0.70 \\ \hline

\multicolumn{1}{c|}{StereoSplat} &
\multicolumn{1}{c|}{\checkmark} &
\multicolumn{1}{c|}{\checkmark} &
\multicolumn{1}{c|}{\checkmark} &
\multicolumn{1}{c|}{} &
\multicolumn{1}{c|}{24.89/0.88} &
\multicolumn{1}{c|}{20.19/0.67} &
\multicolumn{1}{c|}{18.17/0.57} &
20.76/0.70 \\ \hline

\multicolumn{1}{c|}{StereoSplat+} &
\multicolumn{1}{c|}{\checkmark} &
\multicolumn{1}{c|}{\checkmark} &
\multicolumn{1}{c|}{\checkmark} &
\multicolumn{1}{c|}{\checkmark} &
\multicolumn{1}{c|}{\textcolor{red}{\textbf{26.08/0.89}}} &
\multicolumn{1}{c|}{\textbf{\textcolor{red}{20.48/0.70}}} &
\multicolumn{1}{c|}{\textbf{\textcolor{red}{18.40/0.59}}} &
\textbf{\textcolor{red}{21.21/0.72}} \\ \hline

\end{tabular}

\label{ablation_studies_on_kitti360}
\end{table*}
\begin{figure*}[!t]
\centering
\includegraphics[width=0.95 \linewidth]{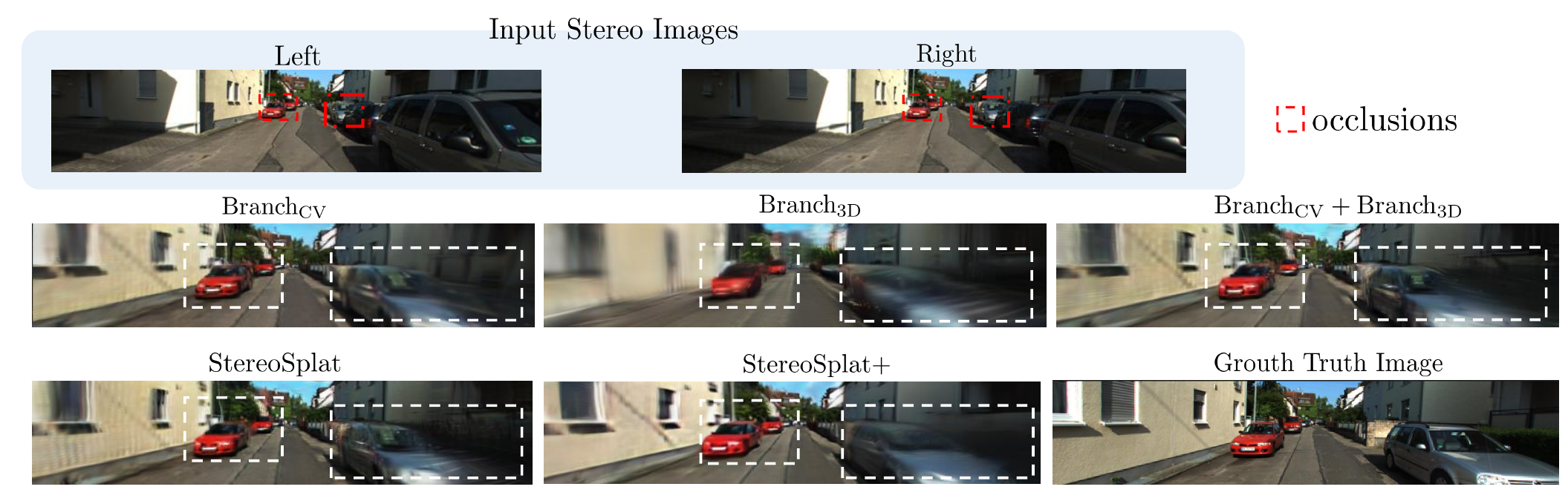}
\caption{Ablation study on rendered forward views. For each bin, we use the first-frame stereo pair as input and render the last-frame right view. The red box indicates occluded regions caused by foreground objects, i.e., parts of vehicles/objects that are not visible in the stereo images due to occlusion. Visual differences are highlighted in the white bounding box.}
\label{fig:ablations:last_view}
\vspace{-2mm}
\end{figure*}

\subsection{Datasets}
\label{sec:exp:datasets}
We evaluate our method on KITTI-360~\cite{kitti360}, an urban-driving benchmark that provides rectified stereo images and sparse LiDAR annotations. Camera poses are obtained from synchronized IMU/GPS, enabling geometry-consistent evaluation. KITTI-360 contains 9 driving sequences; following common practice, we exclude sequences $0003$, $0007$, and $0010$ due to strong scene dynamics that break geometric consistency for 3D Gaussian training, and use the remaining 6 sequences for training and evaluation. As illustrated in Fig.~\ref{fig:dataset_configuration} and following the preprocessing protocol of OmniScene~\cite{OmniScene}, we partition each sequence into trajectory bins, where the distance between the first and last stereo frames is fixed to 8.0 m. Since the stereo rig is forward-moving, we take the first stereo pair in each bin as input, while the remaining frames in the bin (center and last stereo pairs) serve as novel target views for supervision and evaluation. This results in 54{,}862 valid bins in total, which we split 9:1 into 49{,}377 training bins and 5{,}485 evaluation bins. Unless otherwise specified, we train for 1450000 iterations at a resolution of $112\times 544$.

\subsection{Implementation Details}
\label{sec:exp:implementation_details}

\subsubsection{Evaluation Metrics.}
We evaluate novel-view RGB quality using Peak Signal-to-Noise Ratio (PSNR) and Structural Similarity (SSIM), where higher values indicate better fidelity to the reference image.
To \emph{also} quantify geometric accuracy, we report depth errors using Absolute Relative Error (AbsRel) and Squared Relative Error (SqRel), where lower values indicate better depth estimation. Here, $d^{*}$ denotes the sparse LiDAR depth projected into the target camera, and $\Omega$ is the set of valid pixels with LiDAR measurements (after standard validity filtering).
For our method, the evaluated depth $d$ is the depth rendered from the \textit{final} predicted 3DGS using the same camera intrinsics/extrinsics as RGB rendering, ensuring consistency between appearance and geometry.
For each bin, we report PSNR/SSIM and AbsRel/SqRel on the \textit{first frame}, \textit{center frame}, and \textit{last frame}, as well as the \textit{average} over all frames inside bin to comprehensively assess performance across the sequence.

\subsubsection{Experimental Setup.}
We train \textit{StereoSplat} in PyTorch with 4 NVIDIA A6000 GPUs, using AdamW with betas $(0.9,0.99)$, an initial learning rate of $8\times10^{-5}$, and a constant-with-warmup schedule (1k warmup steps). Training is run for 30 epochs with a batch size of 4. The cost-volume branch operates at $1/4$ resolution (downscale factor 4) with a disparity search range of 192. The triplane branch adopts three axis-aligned feature planes with grid size $H\times W\times Z=192\times192\times16$, embedding dimension 128, 8 attention heads, and 3 transformer layers (two cross-view hybrid/image cross-attention layers followed by one self-attention layer). We enable gradient checkpointing to reduce memory usage. In Eq.~\ref{chapter6:eq:loss_function}, we set $\lambda_{\text{1}}=1.0$, $\lambda_{\text{2}} \& \lambda_{\text{3}}=0.01$, and $\lambda_{\text{4}}=0.05$. For image reconstruction, we combine an $\ell_2$ loss and LPIPS with weights $\lambda_{1}=1.0$ and $\lambda_{2}=0.05$, respectively.

\subsection{Ablation Studies}
\label{sec:experiments:ablation_studies}
To verify the effectiveness of each proposed component, training strategy, and the overall StereoSplat+ pipeline, we conduct ablations on the KITTI\textendash360 dataset (seq.~0000). We use 9{,}207 bins for training and the remaining 1{,}023 bins for evaluation. We report all the results in Table~\ref{ablation_studies_on_kitti360}.

\noindent \textbf{Branch Isolation and Fusion.} As shown in Table~\ref{ablation_studies_on_kitti360}, using only the cost\textendash volume branch (Branch$_{\text{CV}}$) peaks on the \textit{first} frame (closest to inputs) thanks to metric disparity cues, but drops on \textit{center/last} frames where occlusion occurs. The triplane-only variant (Branch$_{\text{3D}}$) is more balanced across frames yet lacks the pixel-accurate localization of Branch$_{\text{CV}}$ near the input pose. The fusion version \textit{StereoSplat} (both branches) improves average PSNR/SSIM over either single branch, showing that explicit 3D aggregation complements cost-volume cues in 3D reconstruction. As shown in Figure.~\ref{fig:ablations:last_view}, the cost-volume–only and 3D-volume–only variants fail to model occlusions in the forward-facing novel view, whereas the fused model yields more accurate Gaussian parameters, producing cleaner geometry and texture.

\noindent \textbf{Training Strategy.} Furthermore, with the stochastic stereo view subsampling strategy (Subsection~\ref{method:training:stochastic_stereo_view}), the model achieves more \emph{uniform} improvements across \textit{all} frames, increasing the average PSNR from 20.53 to 20.76. We attribute this gain to the resulting input-invariant view combinations, which encourage the triplane Transformer to learn more robust, view-agnostic features. In this way, stochastic view subsampling better exploits the capacity of the 3D Gaussian estimator under diverse input-view conditions.

\begin{figure*}[!t]
\centering
\includegraphics[width=0.95\linewidth]{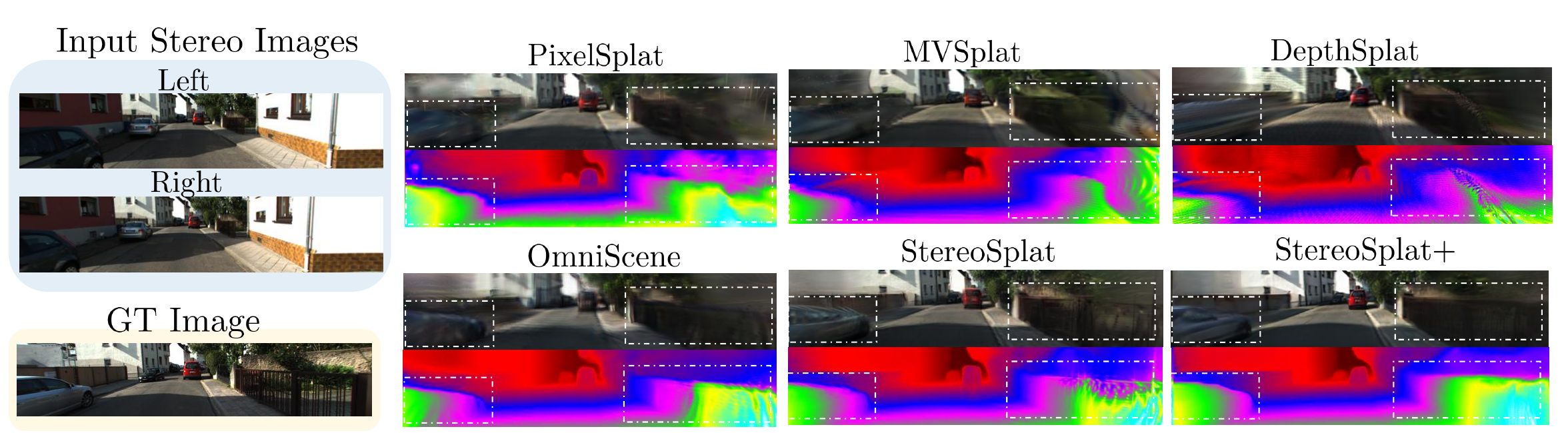}
\caption{Qualitative comparison on KITTI-360 using only the first-frame stereo pair. From the left/right inputs (top left), each method reconstructs 3D Gaussians and renders a novel center view; GT is shown at top right. For each method, we show rendered RGB (row 1) and depth (row 2, jet). White dashed boxes mark challenging regions (thin structures, moving objects).
}
\label{fig:compared_with_othes}
\end{figure*}

\noindent \textbf{Diffusion-Based Progressive Inference.}
Since \textit{StereoSplat} supports a variable number of input stereo pairs, we further investigate a \emph{one-shot progressive} feed-forward schedule (Table~\ref{ablation_studies_on_kitti360}).
Starting from a single stereo pair, we first estimate a 3DGS, render pseudo-novel stereo views, and re-inject these views to increase the effective number of inputs.
However, directly re-injecting rendered views introduces a noticeable domain gap.
Applying our pre-trained one-step diffusion enhancer before re-injection (\textit{StereoSplat+}) substantially mitigates this gap, and we also perform confidence-guided fusion between the initial and refined 3DGS to further improve the quality, achieving the best performance in terms of PSNR of 21.21 and SSIM of 0.72.

\begin{table}[!t]
\centering
\setlength{\tabcolsep}{3.5mm}
\renewcommand{\arraystretch}{0.95} 
\caption{Comparison of feed-forward 3D Gaussian methods on KITTI-360 using a single stereo pair (first frame) as input. We render all frames and report bin-wise average RGB and depth metrics. Best is in \textbf{\textcolor{red}{red}}; second-best is in \textbf{bold}.}
\label{compared_with_others}
\begin{tabular}{ccccc}
\hline
\multirow{3}{*}{\textbf{Method}} & \multicolumn{4}{c}{\textbf{Bin Average}} \\ \cline{2-5}
& \multicolumn{2}{c}{\textbf{RGB}} & \multicolumn{2}{c}{\textbf{Depth}} \\ \cline{2-5}
& \textbf{PSNR} & \textbf{SSIM} & \textbf{AbsRel} & \textbf{SqRel} \\ \hline

\begin{tabular}[c]{@{}c@{}}PixelSplat~\cite{pixelsplat}\end{tabular} 
& 19.04 & 0.60 & \textbf{0.094} & 0.579 \\ \hline

\begin{tabular}[c]{@{}c@{}}MVSplat~\cite{MVSPlaT}\end{tabular} 
& 18.69 & 0.53 & 0.098 & 0.619 \\ \hline

\begin{tabular}[c]{@{}c@{}}OmniScene~\cite{OmniScene}\end{tabular} 
& 18.19 & 0.53 & 0.117 & 0.657 \\ \hline

\begin{tabular}[c]{@{}c@{}}DepthSplat~\cite{DepthSplat}\end{tabular} 
& 18.60 & 0.60 & 0.140 & 0.767 \\ \hline

\textbf{\begin{tabular}[c]{@{}c@{}}StereoSplat\\ (Ours)\end{tabular}} 
& \textbf{20.42} & \textbf{0.67} & \textcolor{red}{\textbf{0.071}} & \textcolor{red}{\textbf{0.386}} \\ \hline

\textbf{\begin{tabular}[c]{@{}c@{}}StereoSplat+\\ (Ours)\end{tabular}} 
& \textbf{\textcolor{red}{20.76}} & \textbf{\textcolor{red}{0.69}} & 0.096 & \textbf{0.487} \\ \hline
\end{tabular}
\end{table}

\subsection{Performance Evaluation}
\label{sec:experiments:performance_evaluation}
To verify the effectiveness of \textit{StereoSplat+}, we compare against recent top-performing feed-forward baselines, including PixelSplat~\cite{pixelsplat}, MVSplat~\cite{MVSPlaT}, OmniScene~\cite{OmniScene}, and DepthSplat~\cite{DepthSplat}, on the full KITTI-360 validation set following the protocol in Subsection~\ref{sec:exp:implementation_details}. We evaluate both appearance and geometry using PSNR/SSIM for novel-view RGB and AbsRel/SqRel for rendered depth. 

As shown in Table~\ref{compared_with_others}, even without diffusion, \textit{StereoSplat} performs strongly on the metric of PSNR and SSIM, indicating high fidelity when rendering near the observed stereo pair. The predicted 3DGS also exhibits improved geometry, reflected by lower rendered-depth errors. After enabling pseudo-view enhancement with diffusion, our \textit{StereoSplat+} further improves the \emph{average} quality to 20.76/0.69, surpassing the strongest prior baseline PixelSplat (19.04/0.60) by +1.72\,dB and +0.09 SSIM. Overall, these results suggest that (i) our feed-forward backbone preserves high-quality renderings near the input views, and (ii) a single diffusion-enhanced enables plug-and-play performance gains for input-invariant multi-view inference.

\begin{figure}[!t]
\centering
\includegraphics[width=1.0\linewidth]{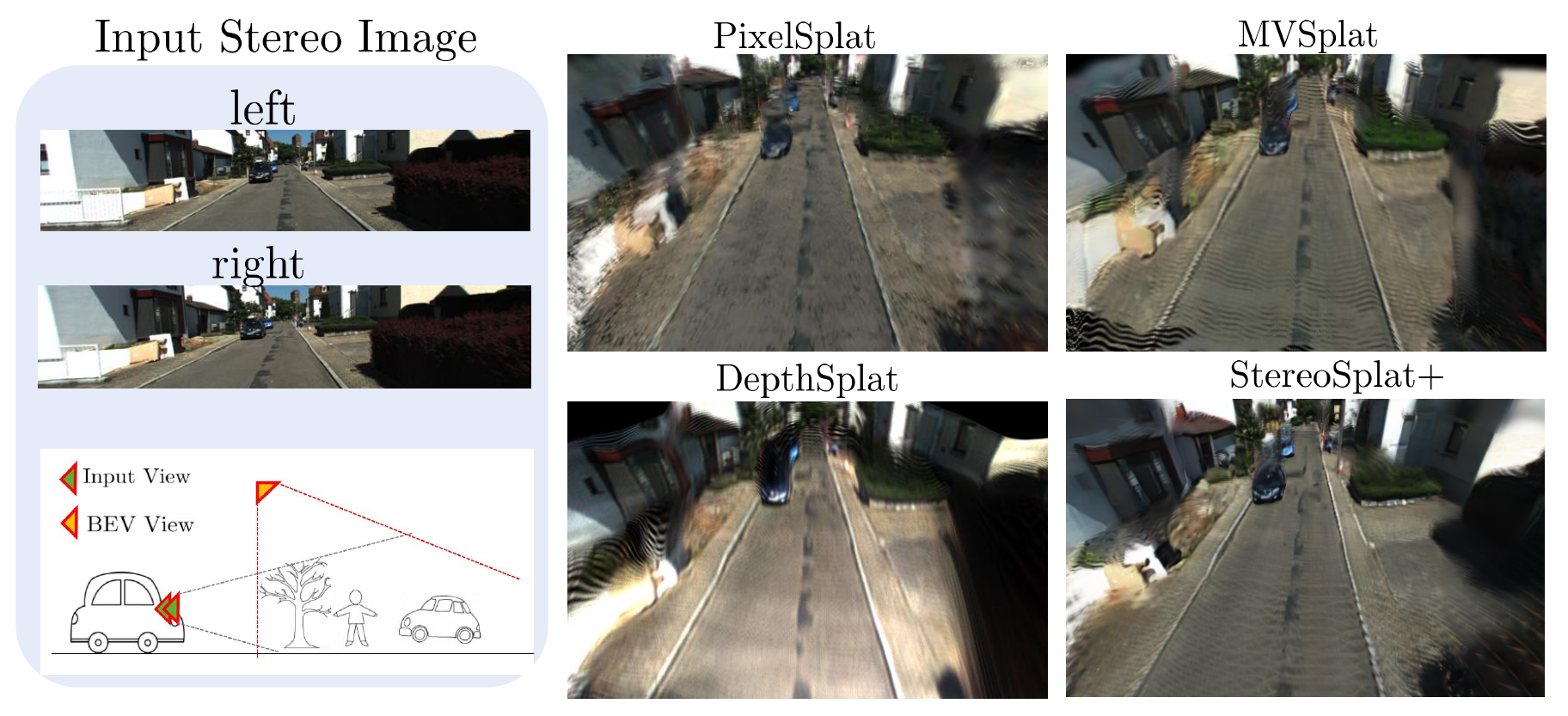}
\caption{\textbf{Bird’s-eye-view extrapolation from only the first-frame stereo pair on KITTI-360.}
Given the rectified stereo input (left), we reconstruct a 3D Gaussian scene and render a distant bird’s-eye view (bottom-left) using strong viewpoint extrapolation. Existing methods produce severe ripples and distorted road/building geometry under this large out-of-plane extrapolation, whereas our StereoSplat+ yields a much more stable BEV image with straighter lanes and cleaner façades.}
\label{fig:compared_with_othes_bev}
\vspace{-5mm}
\end{figure}

As shown in Figure~\ref{fig:compared_with_othes}, existing feed-forward 3D Gaussian methods struggle to reconstruct a stable and geometrically consistent scene from only the first-frame stereo pair by exhibiting motion blur and ghosting around the moving cars and buildings, and showing over-smooth depth, distorting thin structures, and producing erroneous road boundaries, as highlighted by the dash boxes. In contrast, StereoSplat already produces sharper RGB renderings and more coherent depth, and the diffusion-enhanced \textit{StereoSplat+} further suppresses floaters and ringing artifacts. Besides, as shown in Figure~\ref{fig:compared_with_othes_bev}, extrapolating from the first-frame stereo pair to a distant bird's-eye-view is highly challenging.
PixelSplat, MVSplat, and DepthSplat suffer from severe ripple artifacts and distorted road/facade geometry under large out-of-plane viewpoint changes.
In contrast, our StereoSplat+ (one-shot prog.) reconstructs a more coherent 3D scene and therefore produces a significantly cleaner BEV rendering, with straighter lane markings and more stable building structures under such strong extrapolation.

\begin{figure}[!t]
\centering
\includegraphics[width=1.0\linewidth]{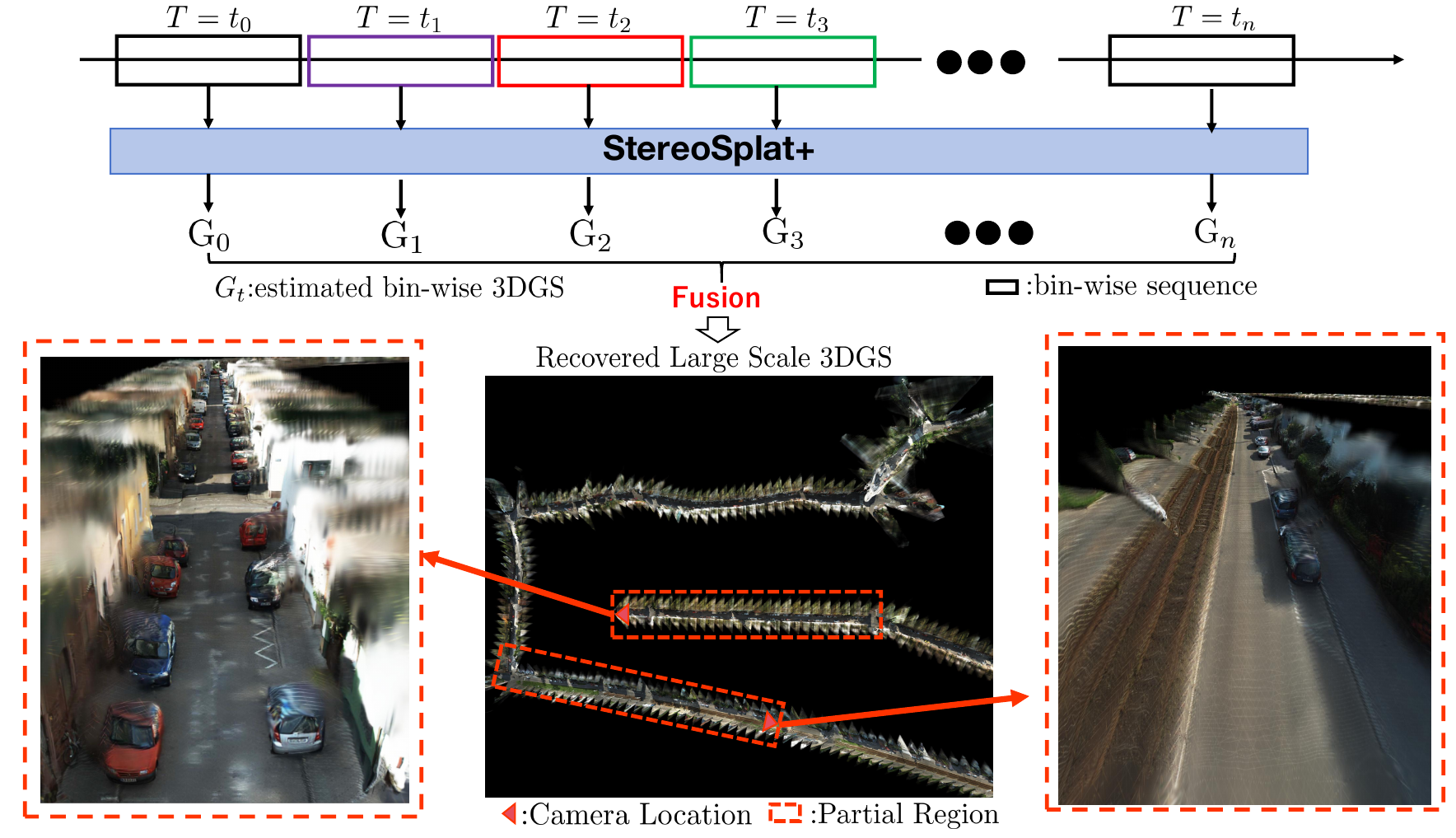}
\caption{\textbf{Incremental large-scale 3DGS reconstruction with known poses.}
Given a long stereo sequence with camera poses, we partition frames into temporal bins and apply \textit{StereoSplat+} to each bin to estimate a bin-wise 3D Gaussian map $G_t$.
We then fuse the resulting Gaussian sets $\{G_t\}$ in the global coordinate frame to recover a consistent global 3DGS map, enabling incremental large-scale reconstruction along long trajectories.}
\label{fig:fusion}
\end{figure}

Besides, in Figure~\ref{fig:fusion}, we further show one potential usage of \textit{StereoSplat+} by leveraging known camera poses to fuse the estimated bin-wise 3D Gaussian maps $\{G_t\}$ into a single global map, enabling incremental large-scale 3D reconstruction. Please refer to \textit{supplementary video} for more details.

\section{Limitations}
Since our proposed \textit{StereoSplat+} relies on a one-step diffusion enhancer, which may occasionally introduce artifacts and can accumulate errors over multiple progressive updates. Moreover, we do not explicitly model dynamic objects, which can cause temporal inconsistency when performing incremental bin-wise fusion for large-scale reconstruction.
\label{sec:limitation}

\section{Conclusions}
\label{sec:conclusions}

In this paper, we introduce \textbf{StereoSplat+}, a progressive, input-invariant, feed-forward 3D Gaussian Splatting (3DGS) framework for stereo inputs. Our \textit{StereoSplat} backbone combines a cost-volume branch with a 3D volume branch to jointly estimate depth and 3D Gaussians, while remaining robust to varying numbers and configurations of input stereo views. To enable progressive inference, we employ a pre-trained one-step diffusion enhancer that transforms noisy renders into reliable pseudo views, allowing self-augmentation that improves coverage and geometry without test-time optimization. 
Experiments on KITTI-360 demonstrate that \textbf{StereoSplat+} consistently outperforms prior feed-forward 3DGS methods in both novel-view synthesis and depth estimation, with particularly strong gains in occluded regions and under large viewpoint extrapolation.

\bibliographystyle{IEEEtran}
\bibliography{IEEEexample}

\end{document}